\documentclass[letterpaper]{article} 
\usepackage{aaai25}  
\usepackage{times}  
\usepackage{helvet}  
\usepackage{courier}  
\usepackage[hyphens]{url}  
\usepackage{graphicx} 
\usepackage{natbib}  
\usepackage{caption} 
\usepackage{algorithm}
\usepackage{algorithmic}

\usepackage{amsmath} 
\usepackage{amssymb}

\usepackage{microtype}
\usepackage{epsfig}
\usepackage{placeins}

\usepackage{enumitem}
\usepackage{tabularx}
\usepackage{xstring}
\usepackage{xspace}
\usepackage{subcaption}       
\usepackage[dvipsnames]{xcolor}
\usepackage{bm}
\usepackage{booktabs}
\usepackage{multirow}
\usepackage{color, colortbl}
\usepackage{epsfig}

\makeatletter
\DeclareRobustCommand\onedot{\futurelet\@let@token\@onedot}
\def\@onedot{\ifx\@let@token.\else.\null\fi\xspace}

\def\eg{\emph{e.g}\onedot} 

\def\ie{\emph{i.e}\onedot}

\def\etc{\emph{etc}\onedot}

\makeatother

\definecolor{gain}{HTML}{34a853}  %
\newcommand{\gain}[1]{\textcolor{gain}{#1}}
\definecolor{lost}{HTML}{ea4335}  %

\newcommand{\res}[2]{{#1} {({\footnotesize{\gain{#2}}})}}

\definecolor{gtable}{rgb}{0.0, 0.5, 0.0}

\newcommand{\ra}[1]{\renewcommand{\arraystretch}{#1}}
\makeatletter

\newcommand{\Rmnum}[1]{\expandafter\@slowromancap\romannumeral #1@}
\makeatother

\usepackage{newfloat}
\usepackage{listings}
\DeclareCaptionStyle{ruled}{labelfont=normalfont,labelsep=colon,strut=off} 
\floatstyle{ruled}
\newfloat{listing}{tb}{lst}{}
\floatname{listing}{Listing}
\title{AdaDiff: Adaptive Step Selection for Fast Diffusion Models}
\author{
    Hui Zhang\textsuperscript{\rm 1,2,3}
    Zuxuan Wu\textsuperscript{\rm 1,2,}\thanks{Corresponding author.},
    Zhen Xing\textsuperscript{\rm 1,2},
    Jie Shao\textsuperscript{\rm 3},
    Yu-Gang Jiang\textsuperscript{\rm 1,2}
}
\affiliations{
    \textsuperscript{\rm 1}Shanghai Key Lab of Intell. Info. Processing, School of CS, Fudan University\\
    \textsuperscript{\rm 2}Shanghai Collaborative Innovation Center of Intelligent Visual Computing\\
    \textsuperscript{\rm 3}ByteDance Inc.
}

\begin{document}

\maketitle

\begin{abstract}
Diffusion models, as a type of generative model, have achieved impressive results in generating images and videos conditioned on textual conditions. However, the generation process of diffusion models involves denoising dozens of steps to produce photorealistic images/videos, which is computationally expensive. Unlike previous methods that design ``one-size-fits-all'' approaches for speed up, we argue denoising steps should be sample-specific conditioned on the richness of input texts. To this end, we introduce AdaDiff, a lightweight framework designed to learn instance-specific step usage policies, which are then used by the diffusion model for generation. AdaDiff is optimized using a policy gradient method to maximize a carefully designed reward function, balancing inference time and generation quality. We conduct experiments on three image generation and two video generation benchmarks and demonstrate that our approach achieves similar visual quality compared to the baseline using a fixed 50 denoising steps while reducing inference time by at least 33\%, going as high as 40\%.
Furthermore, our method can be used on top of other acceleration methods to provide further speed benefits.
Lastly, qualitative analysis shows that AdaDiff allocates more steps to more informative prompts and fewer steps to simpler prompts.
\end{abstract}

%

\section{Introduction}
Diffusion models~\cite{podell2023sdxl,sora}, as a class of generative models, have made significant strides. These models have the capability to generate specified visual content, including images and videos, based on specific input conditions such as text, semantic maps, representations, and images.
For example, models like SDXL~\cite{podell2023sdxl} and Sora~\cite{sora} can produce perceptually-convincing or artistic images and videos conditioned on textual descriptions, \emph{a.k.a.} prompts. These diffusion models often contain an iterative denoising process during generation, and more iterations typically indicate better visual quality. However, the improvements come at the cost of increased computational resources, even with the use of state-of-the-art sampling methods~\cite{song2021ddim,lu2022dpm}. Therefore, to strike a balance between quality and inference speed, the number of denoising steps is often empirically set as a fixed value.

\begin{figure}[h]
  \centering
  \includegraphics[width=0.9\linewidth]{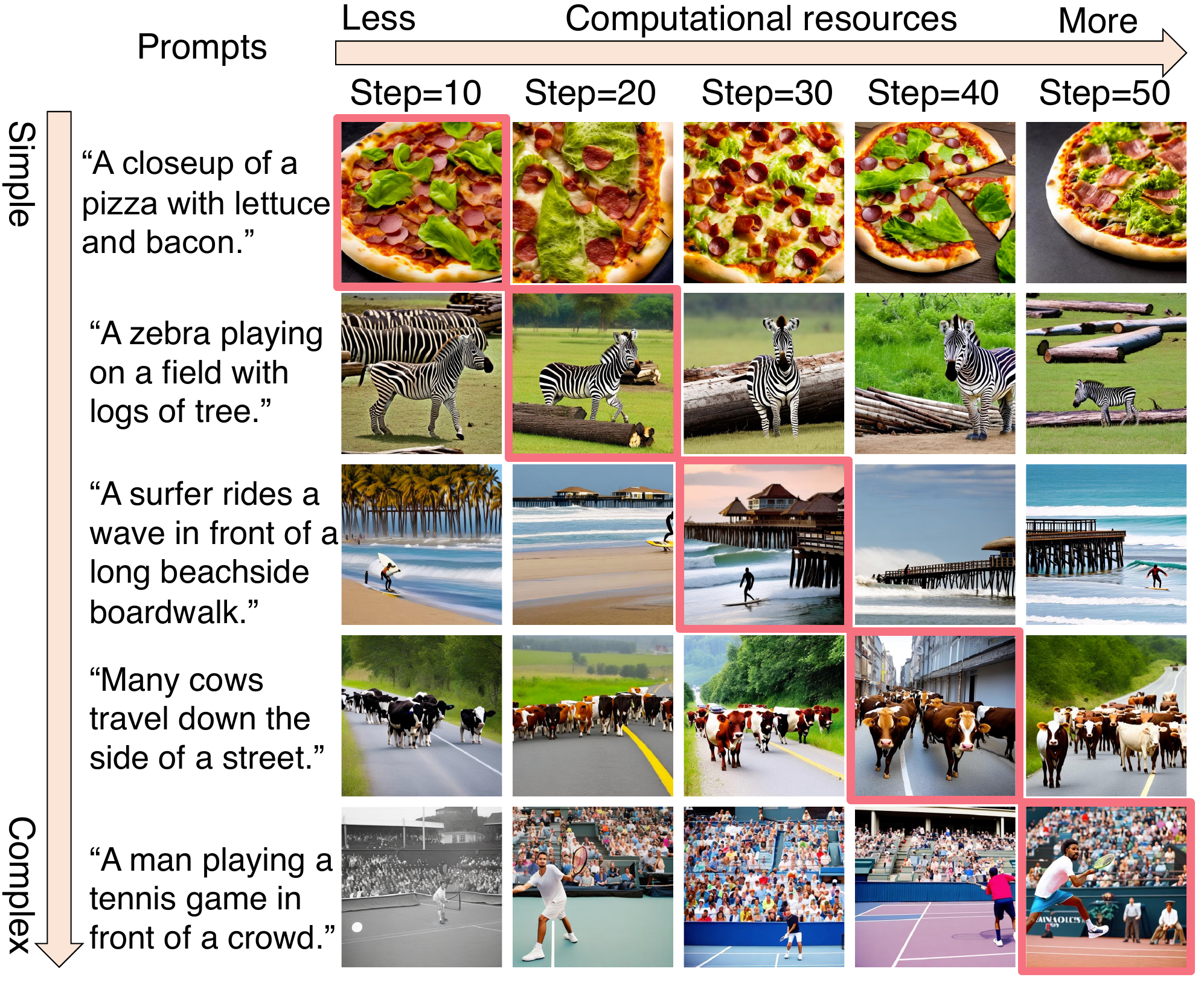}
  \caption{\textbf{A conceptual overview of our approach.} AdaDiff assigns instance-specific denoising steps based on prompt richness to minimize inference time with minimal quality loss. Images with red borders are produced by AdaDiff.}
  \label{fig:teaser}
\end{figure}

\emph{But do we really need a fixed number of denoising steps for all different prompts? }
Intuitively, using more steps may lead to higher quality and more detailed content. 
Nonetheless, in real-world applications, the richness in textual prompts, \ie, the number of objects, and how they relate to each other, vary significantly. For certain easy and coarse-grained prompts, involving only one or a few objects, using fewer steps is already sufficient to generate satisfactory results, and increasing the number of steps may lead to only marginal improvements and does not necessarily produce better results.
For complex textual prompts, containing many objects, detailed descriptions, and intricate interactions between objects, a larger number of steps becomes necessary to achieve the desired result.
Therefore, the goal of this paper is to develop a dynamic framework for diffusion models by adaptively determining the number of denoising steps needed for generating photorealistic contents conditioned on textual inputs. 
This is \emph{in contrast yet complimentary} to existing methodologies, which can be categorized into two main types: \Rmnum{1}) reducing the number of steps via the faster schedulers~\cite{lu2022dpm} or step distillation~\cite{sauer2023sdxl-turbo}; \Rmnum{2}) reducing the computation per step through model pruning and distillation~\cite{SSD-1B}, or by mitigating redundant computation~\cite{ma2024deepcache}. While these methods achieve notable acceleration gains, they typically adopt a ``one-size-fits-all'' strategy, \ie applying the same number of steps without considering the rich complexity of textual prompts.

In light of this, we introduce AdaDiff, an end-to-end framework that aims to achieve efficient diffusion models by learning adaptive step selection in the denoising process based on prompts.
For each prompt, deriving a dynamic generation strategy involves: 
\Rmnum{1}) determining the required number of steps for generation and 
\Rmnum{2}) ensuring high-quality generation even with a relatively smaller number of steps.
With this, AdaDiff can allocate more computational resources to more descriptive prompts while using fewer resources for simpler ones.  While this approach is highly appealing, learning the dynamic step selection is a non-trivial task, as it involves non-differentiable decision-making processes.

To address this challenge, AdaDiff is built upon a reinforcement learning framework. 
Specifically, given a prompt, AdaDiff trains a lightweight step selection network to produce a policy for step usage. Subsequently, based on this derived policy, a dynamic sampling process is performed on a pre-trained diffusion model for efficient generation. The step selection network is optimized using a policy gradient method to maximize a meticulously crafted reward function. The primary objective of this reward function is to encourage the generation of high-quality visual content while minimizing computational resources. It is also worth pointing out that the step selection network conditioned on textual inputs is lightweight with negligible computational overhead.

We conduct extensive experiments to evaluate our proposed method, and the results demonstrate that AdaDiff saves between 33\% and 40\% of inference time compared to the baseline using a fixed denoising step while maintaining similar visual quality across various image and video generation benchmarks~\cite{lin2014coco,christoph2022laioncoco,wang2022diffusiondb,xu2016msr-vtt, wang2023internvid}. 
In addition, we demonstrate that our approach can be combined with various existing acceleration paradigms~\cite{sauer2023sdxl-turbo,SSD-1B}. 
Moreover, the learned policy from one dataset can be successfully transferred to another.
Finally, through qualitative and quantitative analysis, we show that AdaDiff flexibly allocates fewer steps for less informative prompts and more for informative prompts.

\section{Related Work}
\label{sec:related work}
\noindent\textbf{Diffusion Models.}
Diffusion models~\cite{ho2020ddpm,dhariwal2021diffusionbeatgans} have emerged as a powerful force in the field of deep generative models, achieving top-notch performance in various applications, spanning image generation~\cite{rombach2022stablediffusion,podell2023sdxl,li2024playground,chen2023pixart,esser2024sd3,li2024hunyuan}, video generation~\cite{wang2023modelscope,blattmann2023svd,sora,bao2024vidu}, and image restoration~\cite{xia2023diffir,gao2023idm}, among others. Notably, in image and video generation, these diffusion models have demonstrated the ability to produce desired results based on diverse input conditions, including text, semantic maps, representations, and images. However, the inherent iterative nature of the diffusion process has led to a substantial demand for computational resources and inference time during the generation process.

\noindent\textbf{Reinforcement Learning in Diffusion Models.} Pre-training objectives of generative models often do not align perfectly with human intent. Therefore, some work focuses on fine-tuning generative models through reinforcement learning~\cite{sutton2018reinforcement} to align their outputs with human preferences, using human feedback or carefully designed reward functions. Typically, these models~\cite{xu2023imagereward,black2023ddpo,fan2023dpok,wallace2023end,wu2023human} enhance aspects such as text-to-image alignment, aesthetic quality, and human-perceived image quality. Nevertheless, we are the first to leverage reinforcement learning to accelerate image and video generation by learning an instance-specific step usage policy.

\noindent
\textbf{Acceleration of Diffusion Models.}
Recently, effort has been made to accelerate the reverse process of diffusion models. These approaches can be broadly categorized into two paradigms: reducing the number of sampling steps and reducing the computation per step. 
The first paradigm focus on designing fewer-step samplers that extract subsequences from the original sequence~\cite{song2021ddim,lu2022dpm,zhao2023unipc} or use knowledge distillation, where a student model learns to approximate a teacher model's output in fewer steps~\cite{meng2023ondistillation,sauer2023sdxl-turbo,luo2023lcm,lin2024sdxl-lightning,ren2024hypersd}.
The second paradigm primarily focuses on model pruning and distillation~\cite{kim2023bksdm,fang2024diff-pruning,SSD-1B} or reducing redundant computations during the denoising process~\cite{ma2024deepcache,wimbauer2024cacheme,li2023fasterdiffusion}.
The proposed AdaDiff method is \emph{orthogonal} to, yet \emph{complementary} to, the aforementioned acceleration methods. While previous methods reduce the total number of sampling steps, they still rely on a manually set and ``one-size-fits-all'' denoising step, ignoring the complexity in textual prompts that are used to condition the generation process. Instead, AdaDiff aims to dynamically decide the optimal step that achieves a balance between visual quality and inference time on a per-input basis, which can be combined with existing speed-up methods.

\section{Methodology}
\label{sec:method}

\begin{figure*}[ht]
  \centering
  \includegraphics[width=0.8\linewidth]{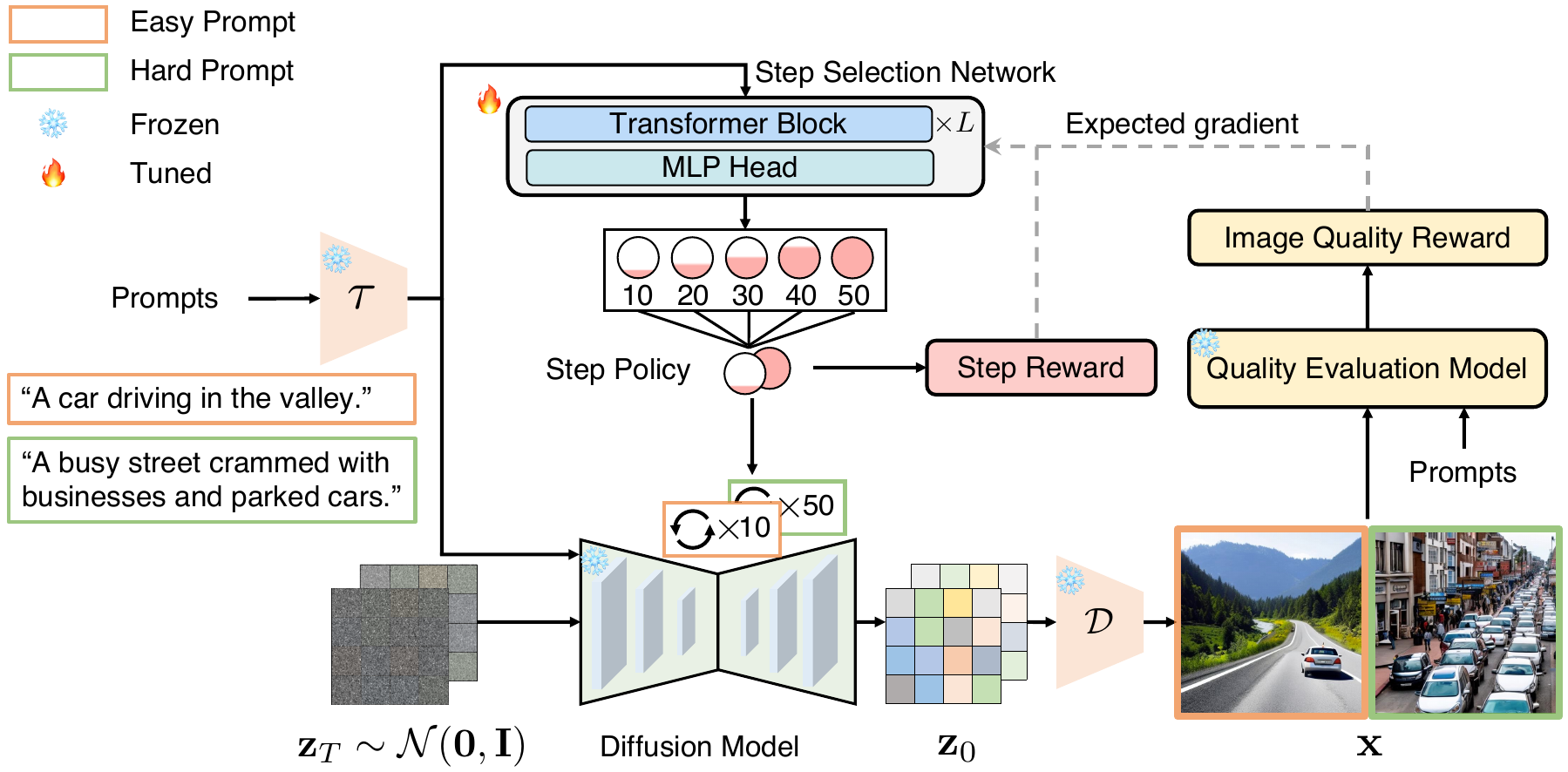}
  \caption{\textbf{An overview of AdaDiff}. Given the input prompts, the step selection network learns the information richness of each prompt and derives the corresponding step usage policy. These policies determine the number of steps required for the diffusion model to generate images. Subsequently, the reward function balances the trade-off between speed and image quality.
  }
  \label{fig:architecture}
\end{figure*}

AdaDiff reduces computational cost and inference time for diffusion models by learning step usage policies conditioned on prompts. The intuition is to encourage using fewer denoising steps while generating high-quality results. In the following, we will first review the background knowledge of diffusion models. 
Subsequently, we will delve into the components of AdaDiff and how it facilitates adaptive generation, including image and video generation.

\subsection{Background on diffusion models}
\label{subsec:preliminaries}
Diffusion modes achieve new state-of-the-art performance in the field of deep generative models inspired by the principles of equilibrium thermodynamics. Specifically, diffusion models involve a forward process where noise is gradually added to the input and a reverse process that learns to recover the desired noise-free data from noisy data. In the forward process, the posterior probability of the diffusion image $x_t$ at time step $t$ has a closed form:
\begin{equation}
\begin{aligned}
    q(x_t|x_0)=\mathcal{N}(x_t;\sqrt{\bar{\alpha_t}}x_0,(1-\bar{\alpha_t})\mathbf{I}),
\end{aligned}
\label{eq:forward}
\end{equation}
where $\bar{\alpha}_{t}=\prod_{i=0}^{t} \alpha_{i}=\prod_{i=0}^{t}(1-\beta_{i})$ and $\beta_{i} \in(0,1)$ represents the noise variance schedule.
Once the diffusion model $\epsilon_{\theta}(x_{t})$ is trained, during the reverse process, traditional diffusion models like DDPM~\cite{ho2020ddpm} denoise $x_{T}\sim\mathcal{N}(\mathbf{0},\mathbf{I})$ step by step for a total of $T$ steps. 
One can also use a discrete-time DDIM~\cite{song2021ddim} sampler to speed up the sampling process.
Initially, the total number of sampling steps $S$ is predetermined. 
Sampling updates are performed at every $\lceil T/S\rceil $ steps according to the plan, reducing the original $T$ steps to a new sampling plan that consists of a subset of $S$ diffusion steps $\hat{T}=\{\eta_{1},\ldots,\eta_{S}\}$. 
DDIM sampler significantly reduces the number of sampling steps and is widely employed in various generative tasks.

\noindent
\textbf{Latent Diffusion Models (LDMs).} LDMs employ an approach where the diffusion process operates in the latent space rather than the pixel space. 
This reduces the training cost and improves the inference speed. 
It uses the pre-trained encoder $\mathcal{E}$ and decoder $\mathcal{D}$ to encode the pixel-space image into a low-dimensional latent and decode the latent back into the image, respectively.
In addition, LDMs incorporate flexible conditional information, such as text conditions and semantic maps, through a cross-attention mechanism to guide the visual generation process. For instance, Stable Diffusion and ModelScopeT2V have been influential approaches for image and video generation conditioned on texts. 
They utilize a DDIM sampler during generation, with a default of 50 steps for all prompts. 
In this paper, AdaDiff aims to enhance the inference speed of LDMs by implementing dynamic step selection on a per-input basis.

\subsection{Adaptive step selection for image generation}
\label{subsec: adaptive image generation}
In image generation, AdaDiff learns a step usage policy conditioned on the text description to reduce the denoising step of Stable Diffusion such that steps vary for different prompts. To this end, AdaDiff builds upon a lightweight selection network trained to determine the total number of steps $\mathbf{t}$ of the DDIM sampler used in the reverse process of Stable Diffusion, as shown in \ref{fig:architecture}. The selection network makes non-differentiable categorical decisions, \ie, steps to be used.

We design $N$ distinct schedulers for the DDIM sampler, each corresponding to a specific total number of sampling steps. 
In this paper, unless specified otherwise, we set $N=10$, which corresponds to the set of step values $ \mathcal{S} =\{5, 10, 15, 20, 25, 30, 35, 40, 45, 50\}$.  
The state space is defined as the input prompts, and actions in the model involve categorizing them in these discrete action spaces.
Then, a carefully designed reward function balances the quality of the generated images with the computational cost.
Formally, given a prompt $\mathbf{p}$, Stable Diffusion first uses a text encoder $\bm{\tau}$ to extract the text features, denoted as $\mathbf{c} = \bm{\tau}(\mathbf{p})$. 
Following this, the step selection network $f_s$, parameterized by $\mathbf{w}$, learns the informativeness of $\mathbf{c}$ using self-attention mechanisms and then further maps it to $\mathbf{s}\in\mathbb{R}^{N}$ through a Multi-Layer Perceptron (MLP):
\begin{equation}
\begin{aligned}
\mathbf{s}=f_s(\mathbf{c}; \mathbf{w}),
\end{aligned}
\label{eq:selection forward}
\end{equation}
where each entity in $\mathbf{s}$ indicates the probability score of choosing this step.
We then define a step selection policy $\pi^{f}(\mathbf{u} \mid \mathbf{p})$ with a $N$-dimensional Categorical Distribution. Here, $\mathbf{u}$ is a one-hot vector of length $N$, denoted as $\mathbf{u}\in\{0,1\}^{N}$, and $\mathbf{u}_{j}=1$ indicates that the step $\mathbf{t}$ with index $j$ in $\mathcal{S}$ is selected.
During training, $\mathbf{u}$ is sampled from the corresponding policy, and during testing, a greedy approach is employed.

So far, the total number of steps $\mathbf{t}$ for DDIM sampling is determined on a per-prompt $\mathbf{p}$ basis. 
Then, Stable Diffusion starts with a latent $\mathbf{z}_T\sim\mathcal{N}(\mathbf{0},\mathbf{I})$, fuses the text-condition features $\mathbf{c}$ through cross-attention, and further generates the desired latent $\mathbf{z}_{0}$ through $\mathbf{t}$ denoising steps. Finally, the decoder $\mathcal{D}$ decodes the latent into a pixel-space generated image $\mathbf{x}$. 
This process is formalized as:
\begin{equation}
\begin{aligned}
\mathbf{x} = \mathcal{D}({LDM}_{sampler}^{\mathbf{p} \rightarrow \mathbf{t}}(\mathbf{z}_T,\mathbf{c})).
\end{aligned}
\label{eq:sampler}
\end{equation}

Recall that the primary objective of AdaDiff is to generate high-quality images in a smaller number of sampling steps, hence, it is crucial to design a suitable reward function to evaluate these actions efficiently.
The reward function consists primarily of two parts: an image quality reward and a step reward, balancing quality and inference time. 
Firstly, to assess image quality, we leverage a quality evaluation model $f_q$ specifically designed to evaluate the quality of generated images~\cite{xu2023imagereward}, abbreviated as the IQS model.
This model assesses image quality in two dimensions: image-text alignment and perceptual fidelity. 
Image-text alignment means that the generated image should match the user-provided text conditions, while perceptual fidelity means that the generated image should be faithful to the shape and characteristics of the object rather than being generated chaotically. 
Typically, the higher the IQS score $f_q(\mathbf{x})$, the higher the quality of the generated image $\mathbf{x}$. Thus, in this paper, we design the image quality reward as $\mathcal{Q}(\mathbf{u}) = f_q(\mathbf{x})$.

For the step reward, we define it as $\mathcal{O}(\mathbf{u}) = 1 - \frac{\mathbf{t}}{S_{max}}$, which represents the normalized steps saved relative to the maximum steps in $\mathcal{S}$.
Finally, the overall reward function is formalized as follows: 
\begin{equation}
\begin{aligned}
R(\mathbf{u})=\left\{\begin{array}{ll}
\mathcal{O}(\mathbf{u}) + \lambda \mathcal{Q}(\mathbf{u}) & \text {for high quality image} \\
-\gamma & \text {else }
\end{array}\right.
\end{aligned}
\label{eq:reward function}
\end{equation}
where $\lambda$ is a hyperparameter that controls the effect of image quality reward $\mathcal{Q}(\mathbf{u})$ and $\gamma$ is the penalty imposed on the reward function when the generated image quality is low. { Instead of a straightforward comparison between the image quality score $f_q(\mathbf{x})$ and a predefined threshold (\eg, 0) to discern whether $\mathbf{x}$ is a high-quality image, we design the determination as a relative manner. 
Specifically, for a given prompt, we individually generate an image for each step in the step set $ \mathcal{S}$ and we consider the image quality score to be high if it ranks top $k$ among these ten images (we empirically set $k$ to 3).
At this point, the step selection network can be optimized to maximize the expected reward:
}
\begin{equation}
\begin{aligned}
\max _{\mathbf{w}} \mathcal{L}=\mathbb{E}_{\mathbf{u} \sim \pi_{f} }R(\mathbf{u}).
\end{aligned}
\label{eq:maximize reward}
\end{equation}
In this paper, we use the policy gradient method~\cite{sutton2018reinforcement} to learn the parameters $\mathbf{w}$ for the step selection network. The expected gradient can be derived as follows:
\begin{equation}
\begin{aligned}
\nabla_{\mathbf{w}} \mathcal{L}=\mathbb{E}\left[R(\mathbf{u}) \nabla_{\mathbf{w}} \log \pi^{f}(\mathbf{u} \mid \mathbf{p})\right],
\end{aligned}
\label{eq:loss}
\end{equation}
which is further approximated with Monte-Carlo sampling using mini-batches:
\begin{equation}
\begin{aligned}
\nabla_{\mathbf{w}} \mathcal{L} \approx \frac{1}{B} \sum_{i=1}^{B}\left[R\left(\mathbf{u}_{i}\right) \nabla_{\mathbf{w}} \log \pi^{f}\left(\mathbf{u}_{i} \mid \mathbf{p}_{i}\right)\right].
\end{aligned}
\label{eq:loss mini-batch}
\end{equation}
where $B$ is the total number of prompts in the mini-batch. The gradient is then propagated back to train the step selection network using the Adam optimizer.

Following the aforementioned training process, the selection network learns the step usage policy that strikes a balance between inference time and generation quality. 
During the inference phase, for different prompts, the maximum probability score in $\mathbf{s}$ is used to determine the number of generation steps, enabling dynamic inference.

\subsection{Adaptive step selection for video generation}
\label{subsec: adaptive video generation}
In addition to image generation, the proposed step selection strategy can also be applied to video diffusion models, such as ModelScopeT2V, as the prompts used to guide video generation also have varying levels of richness. 
The overall implementation paradigm is similar to Figure \ref{fig:architecture}.
To assess the quality of the generated videos $\mathcal{V}$, we individually score each frame using the IQS model and subsequently calculate the average of all frames to obtain the video quality reward.
This procedure can be formalized as $\mathcal{Q}(u) = \frac{1}{F} \sum_{i=1}^{F} f_q(\mathcal{V}_i)$. Here, $F$ is the number of frames in the generated video. Then, we calculate the step reward by $\mathcal{O}(\mathbf{u}) = 1 - \frac{\mathbf{t}}{S_{max}}$, and the overall reward is obtained from Eq.\eqref{eq:reward function}.

\section{Experiments}
\label{sec:experiments}
\subsection{Experimental Details}
\label{subsec:Experimental Details}

\noindent
\textbf{Datasets.}
To evaluate the effectiveness and generalizability of our approach, we conduct extensive experiments on three image datasets: MS COCO 2017~\cite{lin2014coco}, Laion-COCO~\cite{christoph2022laioncoco}, DiffusionDB~\cite{wang2022diffusiondb}, and two video datasets: MSR-VTT~\cite{xu2016msr-vtt} and InternVid~\cite{wang2023internvid}.
In MS COCO 2017, our training set consists of $118,287$ textual descriptions, and all $25,014$ text-image pairs from the validation set are employed for testing. 
Regarding Laion-COCO, we randomly select $200K$ textual descriptions for training and $20K$ text-image pairs for testing. 
The partitioning of the training and testing sets for DiffusionDB follows the same paradigm as Laion-COCO.
The training sets for MSR-VTT and InternVid consist of $6,651$ and $24,911$ text descriptions, respectively. The test set for MSR-VTT comprises $2,870$ text-video pairs.

\noindent
\textbf{Evaluation Metrics.}
Following previous work, we assess the quality of generated images or videos using several metrics, including FID~\cite{heusel2017fid}, IS~\cite{salimans2016is}, CLIP Score~\cite{radford2021clip}, NIQE~\cite{mittal2012niqe}, and the recently introduced Image Quality Score (IQS)~\cite{xu2023imagereward}. 
Additionally, we use the average denoising steps and time per image or video generation to measure the inference speed.

\noindent
\textbf{Implementation Details.}
We design the step selection network as a lightweight architecture consisting of three self-attention layers and a multi-layer perceptron. 
For image generation, we use SD-v2.1-base and SDXL-Turbo to generate $512 \times 512$ images, and SDXL-v1.0 for $1024 \times 1024$ images. 
For video generation, we use ModelScopeT2V to generate 16-frame videos at a resolution of $256 \times 256$.
The parameters and computational cost of the step selection network are 25.71M and 1.93 GFLOPs, respectively, which are negligible compared to SD-v2.1-base's 865M and the 35,140 GFLOPs required to generate an image in 50 steps.
We train the step selection network for 200 epochs with a batch size of 256 and use the Adam optimizer with an initial learning rate of $10^{-5}$.  
The training cost ranges from 16 to 80 A100 GPU hours for the adaptive step policy applied to different base models.

\subsection{Main Results}
\label{subsec:main results}
\noindent
\textbf{Performance on Image and Video Generation.}
To validate the effectiveness and general applicability of AdaDiff, we compare it with the following baseline methods:
\begin{itemize} 
    \item \emph{One-size-fits-all}: The base model uses a fixed number of sampling steps for different prompts. We primarily chose two step counts: the default number (typically 50) and a count close to the speed of AdaDiff.
    \item \emph{Random}: Given the step usage policies produced by AdaDiff, we generate random step policies to validate the effectiveness of learned policies.
    \item \emph{Heuristic}: We propose a heuristic step usage policy, allocating more generation steps for prompts containing more words. For example, for words \textless 8, we allocate 10 steps; for 8 $\leq$ words \textless 10, we allocate 20 steps, and so on. 
    \item \emph{Perplexity}: We further propose a step usage strategy, which allocates more steps for prompts with higher perplexity~\cite{jelinek1977perplexity}. For example, for perplexity \textless 20, we allocate 10 steps; for 20 $\leq$ perplexity \textless 40, we allocate 20 steps, and so on. 
\end{itemize}

\begin{table}[h]
\centering
 \ra{1.1}
 \tabcolsep=0.18cm
\scalebox{0.72}{
\begin{tabular}{@{}lccccccc@{}}
\toprule
\multirow{2}{*}{\textbf{COCO}}      & \multicolumn{2}{c}{\textbf{Speed}} & \multicolumn{5}{c}{\textbf{Image Quality}}    \\ \cmidrule(l){2-3} \cmidrule(l){4-8} 
                 &Step$\downarrow$             &Time$\downarrow$            &IQS$\uparrow$ &CLIP$\uparrow$  &IS$\uparrow$ &FID$\downarrow$   &NIQE$\downarrow$  \\ \midrule
SD-v2.1 & 50                  &   2.24             & 0.419    &0.314          &37.48    &22.13              &3.75      \\\midrule
SD-v2.1 & 28                 &   1.28              & 0.362	&0.312	&37.35	&22.50	&3.88      \\
Random                    & 30.03             & 1.41            & 0.354 & 0.313 & 36.85 & 22.50  & 3.88 \\
Heuristic                 & 29.75             & 1.40             & 0.369 & 0.313 & 36.72 & 22.73 & 3.96 \\
Perplexity                & 31.79             & 1.48            & 0.368 & 0.313 & 37.33 & 22.77 & 3.86 \\ \midrule
\rowcolor[HTML]{e9e9e9} 
AdaDiff & 28.61           &  \res{1.35}{$\uparrow$39.7\%}            &\textbf{0.412}       &\textbf{0.314}     &\textbf{37.60}            &\textbf{21.92}      &\textbf{3.76}          \\ \bottomrule 
\end{tabular}}
\caption{Comparison of AdaDiff on image generation.}
\label{tab: results on COCO-2017}
\end{table}

Table \ref{tab: results on COCO-2017} offers a detailed analysis of AdaDiff's performance on the COCO-2017 benchmarks, with results averaged over five independent runs. AdaDiff assigns average sampling step counts of 28.61. Compared to the fixed 50-step SD, AdaDiff achieves similar performance across five image quality metrics while providing a 39.7\% speed increase. At comparable speeds, AdaDiff significantly surpasses the fixed 28-step SD in image quality.
Furthermore, the step usage policy learned by AdaDiff shows advantages over the random policy and manually crafted policies like heuristic and perplexity-based approaches.
These findings validate that AdaDiff efficiently generates instance-specific step usage policies, allocating different sampling step counts per prompt to optimize speed with minimal loss in image quality.

\begin{table}[h]
\centering
\ra{1.1}
\tabcolsep=0.18cm
\scalebox{0.72}{
\begin{tabular}{@{}lccccccc@{}}
\toprule
\multirow{2}{*}{\textbf{MSR-VTT}} & \multicolumn{2}{c}{\textbf{Speed}} & \multicolumn{5}{c}{\textbf{Video Quality}}    \\ \cmidrule(l){2-3} \cmidrule(l){4-8} 
                 &Step$\downarrow$             &Time$\downarrow$            &IQS$\uparrow$ &CLIP$\uparrow$  &IS$\uparrow$ &FID$\downarrow$   &NIQE$\downarrow$  \\ \midrule
ModelScope            & 50                &   21.2              & -0.518 & 0.293 & 18.79 & 44.85 & 6.37 \\\midrule
ModelScope            & 31                 &  13.1               & -0.678 & 0.293 & 18.42 & 46.09 & 6.52 \\
Random                   & 29.98             & 13.5         & -0.723 & 0.293 & 18.22 & 47.41 & 6.75 \\
Heuristic                & 28.19             & 13.2            & -0.709 & 0.293 & 18.32 & 46.45 & 6.59 \\
Perplexity               & 28.86             & 13.3            & -0.685 & 0.294 & 18.39 & 46.01 & 6.65 \\ \midrule
\rowcolor[HTML]{e9e9e9}  
AdaDiff            & 31.14              &  \res{13.6}{$\uparrow$35.8\%}                & \textbf{-0.532} & \textbf{0.294} & \textbf{18.71} & \textbf{45.02} & \textbf{6.42} \\ \bottomrule
\end{tabular}}
\caption{Comparison of AdaDiff on video generation.}
\label{tab:results on video benchmarks}
\end{table}

Table \ref{tab:results on video benchmarks} confirms the effectiveness of AdaDiff for video generation tasks. Compared to ModelScope's fixed 50-step outputs, AdaDiff not only enhances video quality but also reduces the generation time per video by 35.8\% (13.6 vs 21.2). When compared to the fixed 31-step, random, heuristic, and perplexity-based policies, AdaDiff uses similar computational resources but significantly enhances video quality across all metrics. These results demonstrate the efficacy of AdaDiff in video generation and its potential to be applied across various text-conditioned diffusion models.

\begin{table}[h]
\centering
 \ra{1.1}
 \tabcolsep=0.18cm
 
\scalebox{0.72}{
\begin{tabular}{@{}lccccccc@{}}
\toprule
\multirow{2}{*}{\textbf{COCO}}      & \multicolumn{2}{c}{\textbf{Speed}} & \multicolumn{4}{c}{\textbf{Image Quality}}    \\ \cmidrule(l){2-3} \cmidrule(l){4-7} 
                 &Step$\downarrow$             &Time$\downarrow$            &IQS$\uparrow$  &IS$\uparrow$ &FID$\downarrow$   &NIQE$\downarrow$  \\ \midrule
SDXL-Euler     & 50              & 5.59 & 0.692         & 36.43 & 24.13 & 3.83 \\ \midrule
\multicolumn{8}{@{}l}{\textcolor{gray}{Reduce the number of steps}}                                                  \\
SDXL-DPMSolver & 29              & 3.24 & 0.634         & 35.52 & 24.49 & 3.95   \\
SDXL-Lightning & 8               & 0.92 & 0.652         & 36.14 & 27.75 & 3.85   \\ \midrule
\multicolumn{8}{@{}l}{\textcolor{gray}{Reduce the computation per step}}                                                     \\
SSD-1B         & 50              & 3.60  & 0.588         & 35.03 & 28.99 & 3.95   \\
DeepCache(N=2) & 50              & 3.21  & 0.636         & 34.99 & 24.76 & 4.07   \\ \midrule
\rowcolor[HTML]{e9e9e9} 
AdaDiff-SDXL         & 29.16           & \res{3.31}{$\uparrow$40.8\%} & \textbf{0.678}     & \textbf{36.16} & \textbf{24.31} & 3.88 & \\ \bottomrule 
\end{tabular}}
\caption{Comparison with other acceleration methods}
\label{tab: results on SDXL}
\end{table}

\noindent
\textbf{Compared with other acceleration methods.} In Table \ref{tab: results on SDXL}, we compare AdaDiff with two other acceleration paradigms, ensuring a fair comparison by using the same base model and benchmark—SDXL and the COCO-2017 5k validation set. In comparison with the first paradigm, which reduces the number of steps, AdaDiff outperforms the faster DPMSolver sampler in quality at a similar speed. Compared to step distillation methods like SDXL-Lightning, our method achieves better performance with significantly lower training costs (80 vs 1000+ A100 GPU hours). Although AdaDiff uses more average steps, it can be used on top of step distillation methods, which we discuss later. In the comparison with the second paradigm, which focuses on reducing the computation per step, AdaDiff achieves better image quality at similar speeds compared to methods based on model pruning and distillation like SSD-1B, as well as those that save computational efforts such as DeepCache.

\begin{table}[h]
\centering
\ra{1.1}
\tabcolsep=0.18cm
\scalebox{0.72}{
\begin{tabular}{@{}lccccccc@{}}
\toprule
\multirow{2}*{\textbf{COCO}}              & \multicolumn{2}{c}{\textbf{Speed}} & \multicolumn{5}{c}{\textbf{Image Quality}}    \\ \cmidrule(l){2-3} \cmidrule(l){4-8} 
                 &Step$\downarrow$             &Time$\downarrow$            &IQS$\uparrow$ &CLIP$\uparrow$  &IS$\uparrow$ &FID$\downarrow$   &NIQE$\downarrow$  \\ \midrule

SDXL-Turbo                  & 5                & 0.55           & 0.801 & 0.315 & 43.71 & 26.12 & 4.01 \\ \midrule
SDXL-Turbo                 & 2                & 0.24            & 0.754 & 0.313 & 43.37 & 28.42 & 4.17 \\ 
Random                    & 3.01             & 0.35           & 0.761 & 0.313 & 43.39 & 27.27 & 4.13 \\
Heuristic                 & 2.97             & 0.33           & 0.767 & 0.313 & 43.31 & 27.17 & 4.11 \\
Perplexity                & 3.18             & 0.38          & 0.772 & 0.314 & 43.42 & 26.80  & 4.11 \\\midrule
\rowcolor[HTML]{e9e9e9} 
AdaDiff                   & 2.19           &\res{0.26}{$\uparrow$52.7\%}          & \textbf{0.791} & \textbf{0.314} & \textbf{43.68} & \textbf{26.71} &  \textbf{4.06} \\ \bottomrule
\end{tabular}}
\caption{Extension AdaDiff to SDXL-Turbo.}
\label{tab: results on SDXL-Turbo}
\end{table}

\noindent
\textbf{Extension AdaDiff to other acceleration methods.} 
Although previous methods achieve notable acceleration gains, they still adopt a ``one-size-fits-all'' strategy, \ie applying the same number of steps regardless of the complexity of the prompts. In light of this, AdaDiff can be used on top of these methods to achieve further acceleration. Firstly, we apply AdaDiff to the paradigm of reducing the number of steps, \ie SDXL-Turbo. As shown in Table \ref{tab: results on SDXL-Turbo}, compared to the fixed 5-step results, AdaDiff achieves comparable performance with a 52.7\% speed up. Compared to the fixed 2-step, AdaDiff clearly provides better image quality at similar speeds.
Additionally, we apply AdaDiff to the paradigm of reducing the computation per step, such as SSD-1B. As shown in Table \ref{tab: results on SSD-1B}, AdaDiff achieves 41.1\% speed up compared to fixed 50-steps and outperformed the fixed 30-step, random, heuristic, and perplexity policies at similar speeds. These results confirm that AdaDiff is orthogonal to, yet complementary to, previous acceleration methods, offering broad applicability.

\begin{table}[h]
\centering
 \ra{1.1}
 \tabcolsep=0.18cm
\scalebox{0.75}{
\begin{tabular}{@{}lccccccc@{}}
\toprule
\multirow{2}{*}{\textbf{COCO}}      & \multicolumn{2}{c}{\textbf{Speed}} & \multicolumn{5}{c}{\textbf{Image Quality}}    \\ \cmidrule(l){2-3} \cmidrule(l){4-8} 
                 &Step$\downarrow$             &Time$\downarrow$            &IQS$\uparrow$ &CLIP$\uparrow$  &IS$\uparrow$ &FID$\downarrow$   &NIQE$\downarrow$  \\ \midrule
SSD-1B         & 50                       & 3.60  & 0.588   & 0.336  & 35.03  & 28.99  & 3.95  \\ \midrule
SSD-1B         & 29                       & 2.09 & 0.537   & 0.334  & 34.18  & 29.31  & 4.01  \\ 
Random         & 30.09                    & 2.23 & 0.522   & 0.334  & 34.16  & 29.29  & 4.12  \\
Heuristic      & 29.69                    & 2.18 & 0.531   & 0.335  & 34.21  & 29.33  & 4.08  \\
Perplexity     & 30.11                    & 2.26 & 0.538   & 0.335  & 34.25  & 29.28  & 4.06  \\\midrule
\rowcolor[HTML]{e9e9e9} 
AdaDiff & 29.16                    & \res{2.12}{$\uparrow$41.1\%} & \textbf{0.571}   & \textbf{0.336}  & \textbf{34.77}  & \textbf{28.83}  & \textbf{3.99}  \\ \bottomrule
\end{tabular}}
 \caption{Extension AdaDiff to SSD-1B.}
\label{tab: results on SSD-1B}
\end{table}

\begin{table}[h]
\centering
\ra{1.1}
\tabcolsep=0.18cm
\scalebox{0.70}{
\begin{tabular}{@{}lccccccc@{}}
\toprule
                   & \multicolumn{2}{c}{\textbf{Speed}} & \multicolumn{5}{c}{\textbf{Image / Video Quality}} \\ \cmidrule(l){2-3} \cmidrule(l){4-8}
                   &Step$\downarrow$             &Time$\downarrow$            &IQS$\uparrow$ &CLIP$\uparrow$  &IS$\uparrow$ &FID$\downarrow$   &NIQE$\downarrow$  \\ \midrule
\multicolumn{8}{@{}l}{\textbf{COCO 2017} $\rightarrow$ \textbf{Laion-COCO}}      \\
SD   & 50               & 2.27             & 0.350   & 0.319  & 30.71 & 22.08 & 4.58 \\
\rowcolor[HTML]{e9e9e9} 
AdaDiff & 30.50          &\res{1.38}{$\uparrow$39.2\%}          & 0.341   & 0.320   & 30.61 & 22.15 & 4.62\\ \midrule
\multicolumn{8}{@{}l}{\textbf{InternVid} $\rightarrow$ \textbf{MSR-VTT}}      \\
ModelScope      & 50               & 21.20             & -0.518  & 0.293  & 18.79 & 44.85 & 6.37 \\
\rowcolor[HTML]{e9e9e9} 
AdaDiff & 32.23             & \res{14.03}{$\uparrow$33.8\%}            & -0.521  &0.292  & 18.76 & 45.01 & 6.45\\ \bottomrule
\end{tabular}}
\caption{Validation on zero-shot adaptive generation.}
\label{tab: cross dataset}
\end{table}

\noindent
\textbf{Extension to other datasets.}
We also evaluate AdaDiff's ability to generalize its learned step selection policy from one dataset to another, which we refer to as zero-shot generation performance. For image generation, we evaluate the step usage policy derived from COCO 2017 on Laion-COCO, while for video generation, we use the policy of InternVid for validation on MSR-VTT. As shown in Table \ref{tab: cross dataset}, the dynamic strategy saves 39.2\% and 33.8\% of generation time on the two datasets separately compared to the fixed 50-step generation while maintaining comparable generation quality.
These results demonstrate the transferability of step selection strategies trained on large-scale data.

\begin{figure}[h]
  \centering
  \includegraphics[width=0.9\linewidth]{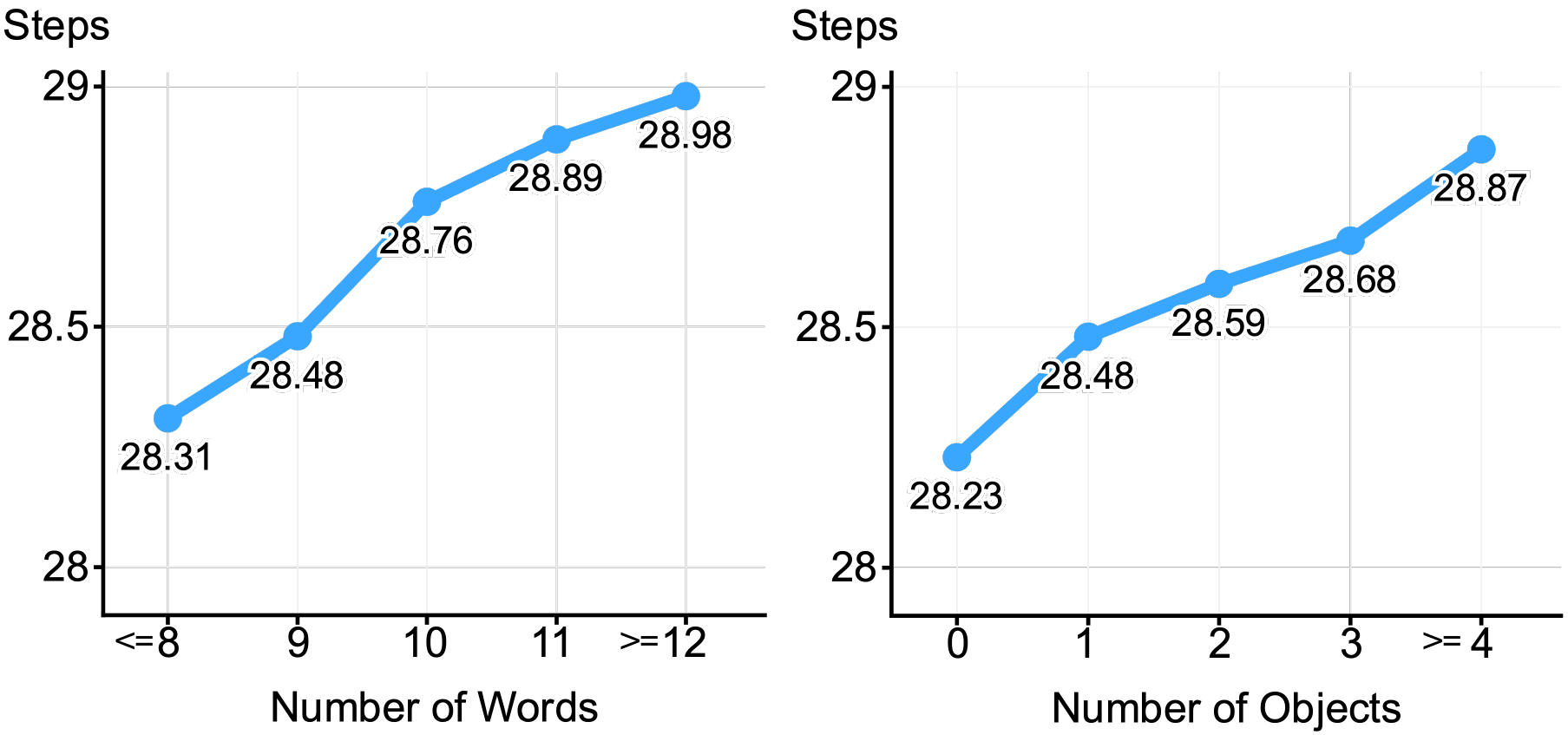}
  \caption{Analyze the learned step policy based on the number of words and objects in the prompts. AdaDiff tends to assign more steps to more informative prompts.}
  \label{fig:quantitative}
\end{figure}

\noindent
\textbf{Analyses of learned policies.}
To better understand the learned policy of AdaDiff, we investigate the relationship between step selection and prompt richness through Figure \ref{fig:quantitative}. 
We evaluate prompt richness in terms of the number of words and objects, assuming an overall increase in information richness as both grow. 
We observe that as the number of words in the prompt increases, the generated results may include more detailed descriptions, spatial relationships, attribute definitions, \etc. 
Consequently, AdaDiff allocates more denoising steps for these richness prompts. 
Besides, as the number of objects in the prompt increases, the results involve a larger amount of details and interactions among the objects. As a result, AdaDiff also assigns more steps for these prompts.

\begin{figure*}[h]
  \centering
  \includegraphics[width=0.85\linewidth]{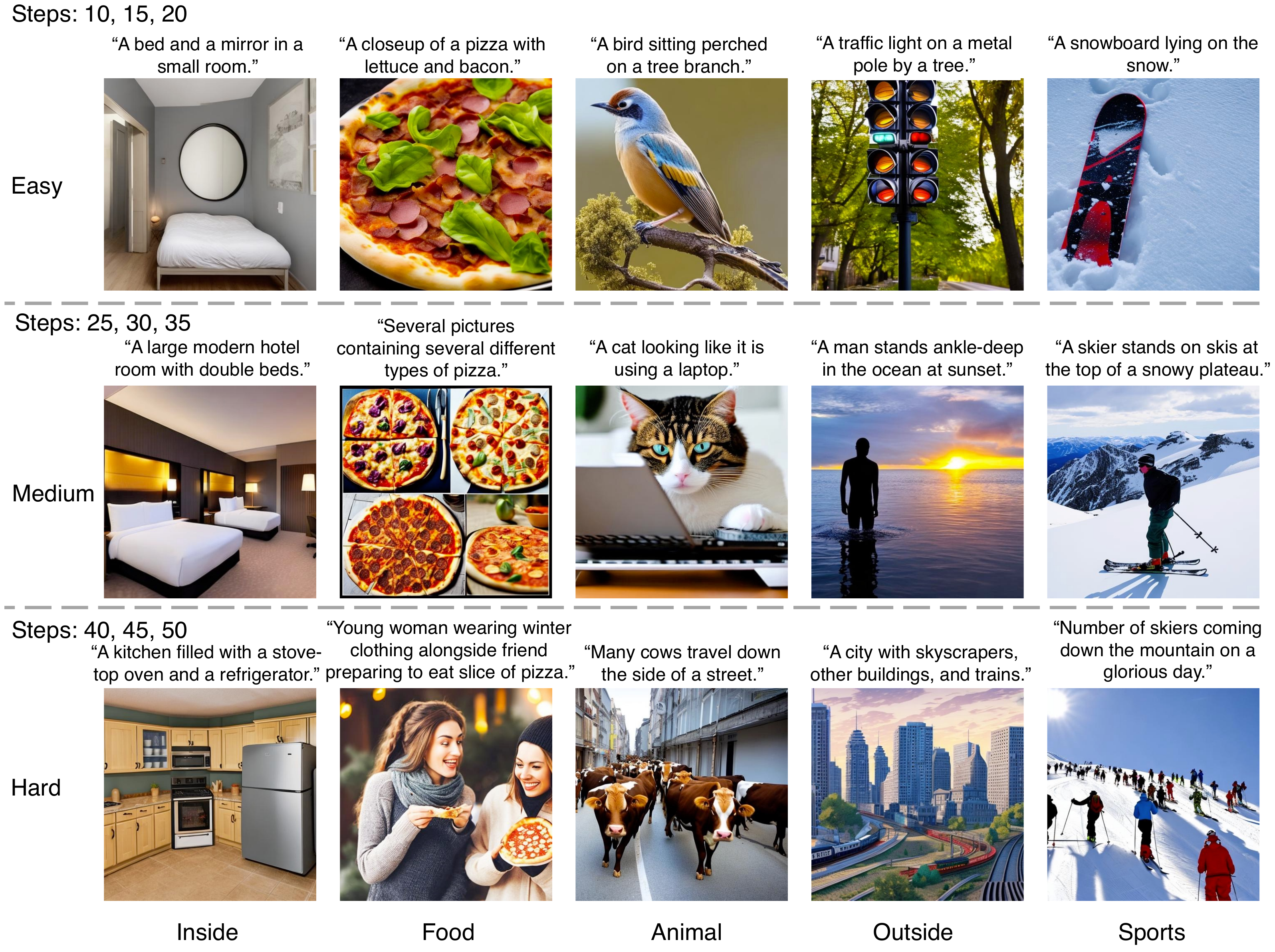}
  \caption{\textbf{Qualitative Results.} AdaDiff implements an instance-specific dynamic generation based on the prompt complexity.
  }
  \label{fig:qualitative}
\end{figure*}

\noindent
\textbf{Qualitative Results.}
We further qualitatively analyze our approach as shown in Figure \ref{fig:qualitative}. We specifically examine five distinct scenarios \{indoor, food, animal, outdoor, and sports\}, and explore the impact of the richness of the prompts on the step usage policy within the same scenario. Our observations reveal that in instances where the prompt is straightforward (easy, Figure \ref{fig:qualitative}), typically involving only one or a few objects, AdaDiff assigns 10 to 20 steps to obtain satisfactory generated results. For prompts characterized by an increased number of objects or the inclusion of detailed descriptions (medium, Figure \ref{fig:qualitative}), such as ``multiple pizzas" or ``ankle-deep", AdaDiff allocates a higher number of steps, typically around 30. In the case of challenging prompts (hard, Figure \ref{fig:qualitative}), which often involve numerous objects, intricate interactions between them, and diverse detailed descriptions, AdaDiff allocates 40 to 50 steps to achieve satisfactory generation.

\subsection{Discussion}
\label{subsec:ablation}
\noindent
\textbf{Reward function.}
The core idea of the reward function is to generate images via fewer steps while maintaining quality.
Table \ref{tab: reward function} showcases various designs of the reward function:
\begin{itemize} 
    \item \emph{Focus only on step savings}: The step selection strategy tends to use fewer steps but ignores quality degradation.
    \item \emph{Integrating step savings with various image quality metrics}: Incorporating image quality metrics such as CLIP, NIQE, and IQS scores into the reward function enhances the step selection strategy's sensitivity to image quality in varying degrees.
    IQS score considers both perceptual fidelity and text-image alignment, significantly improving the strategy's effectiveness. In contrast, focusing solely on text-image alignment (CLIP score) or perceptual fidelity (NIQE score) may lead to quality decline due to the incomplete assessment perspective.
    \item \emph{Different criteria for determining whether an image is high quality}: The relative approach ranks images generated by different steps using IQS scores and defines the top-$k$ images as high quality. In contrast, the absolute approach defines images as high quality if their IQS score exceeds a manually set threshold~(\eg 0). The findings in Table \ref{tab: reward function} indicate that the relative approach learns a more efficient step policy.
\end{itemize}

\begin{table}[h]
\centering
 \ra{1.1}
\tabcolsep=0.18cm
\scalebox{0.71}{
\begin{tabular}{@{}cccccccccc@{}}
\toprule
\multicolumn{6}{c}{\textbf{Reward function}}       & \multicolumn{4}{c}{\textbf{Performance}} \\ \cmidrule(l){1-6} \cmidrule(l){7-10}
Step & CLIP & NIQE & IQS & Top-$k$ & Threshold & Step$\downarrow$  & IQS$\uparrow$    & IS$\uparrow$     & NIQE$\downarrow$  \\ \midrule
$\checkmark$    &               &               &               & $\checkmark$     &          & 18.12  & 0.335  & 36.58  & 3.96 \\
$\checkmark$    & $\checkmark$    &               &                 & $\checkmark$     &          & 18.86  & 0.334  & 36.53  & 3.97 \\
$\checkmark$    &               & $\checkmark$    &                 & $\checkmark$     &          & 29.62  & 0.392  & 37.11  & 3.79 \\
\rowcolor[HTML]{e9e9e9} 
$\checkmark$    &               &                & $\checkmark$   & $\checkmark$     &          & 28.61  & \textbf{0.412}  & \textbf{37.60}  & \textbf{3.76} \\
$\checkmark$    &               &               & $\checkmark$   &                      & $\checkmark$        & 25.23  & 0.377  & 37.25  & 3.91 \\ \bottomrule
\end{tabular}}
\caption{Comparisons of different reward functions.}
\label{tab: reward function}
\end{table}

\begin{figure}[h]
  \centering
  \includegraphics[width=0.9\linewidth]{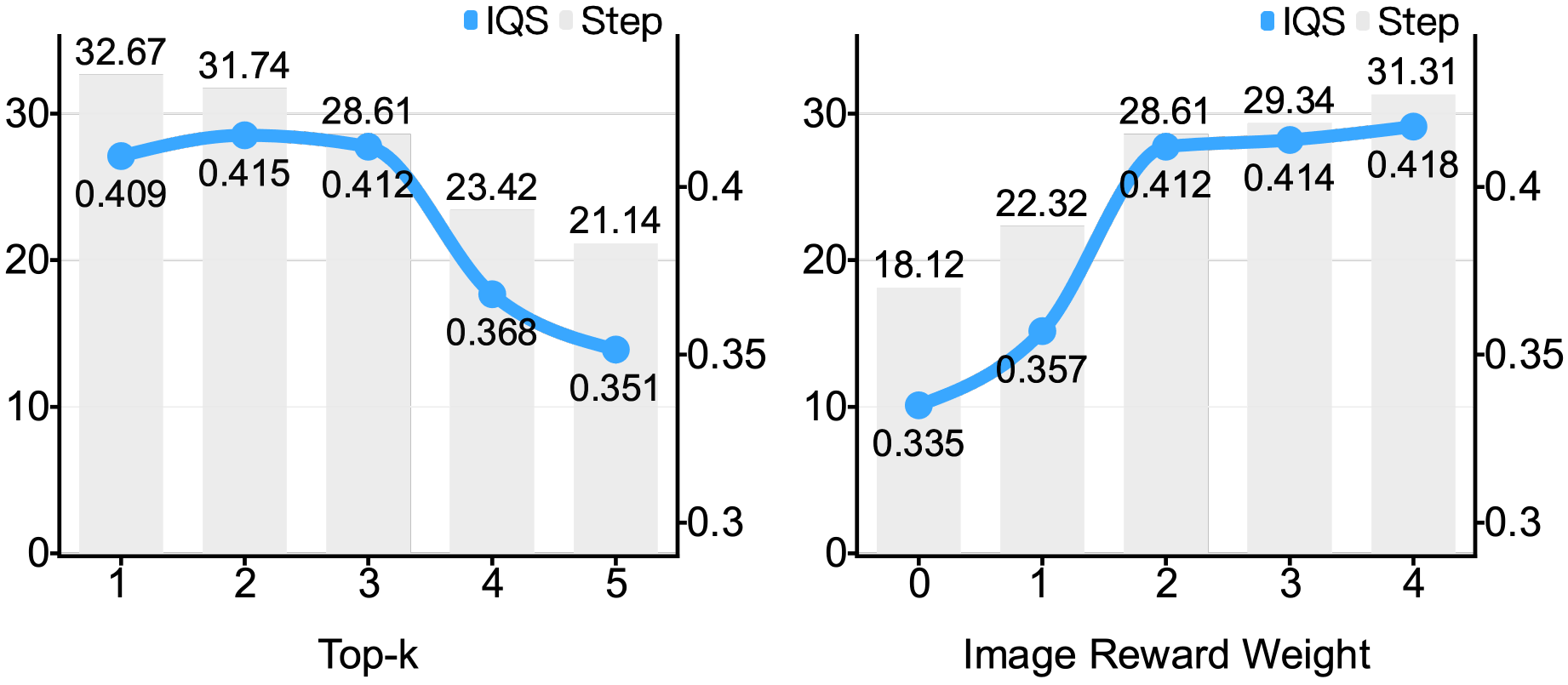}
  \caption{The speed and quality trade-offs modulated by top-$k$ and image reward weight.}
  \label{fig:topk-weight}
\end{figure}

\noindent
\textbf{Different trade-offs between speed and quality.}
The hyperparameters top-$k$ and image reward weight $\lambda$ modulate different speed-quality trade-offs, as shown in Figure \ref{fig:topk-weight}. Smaller values of $k$ and larger values of $\lambda$ enforce higher standards for image quality, resulting in generated images with higher visual quality and better prompt following, accompanied by an increase in the average number of steps.

\section{Conclusion}
\label{sec:conclusion}
In this paper, we introduced AdaDiff, a method that derives adaptive step usage policies tailored to each prompt, facilitating efficient image and video generation. More precisely, a step selection network is trained using policy gradient methods to generate these policies, striking a balance between generation quality and the reduction of overall computational costs. 
Extensive experiments validated the capability of AdaDiff to generate strong step usage policies on per-input bias, providing compelling qualitative and quantitative evidence. Additionally, AdaDiff can be used on top of other acceleration methods to provide further speed benefits.

\section{Acknowledgements}
This project was supported by the National Natural Science Foundation of China under Grant No. 2021ZD0112805 and ByteDance under Grant No. CT20230914000147.

\section{Appendix}

\subsection{More Experimental Details}

\begin{figure}[h]
  \centering
  \includegraphics[width=1.0\linewidth]{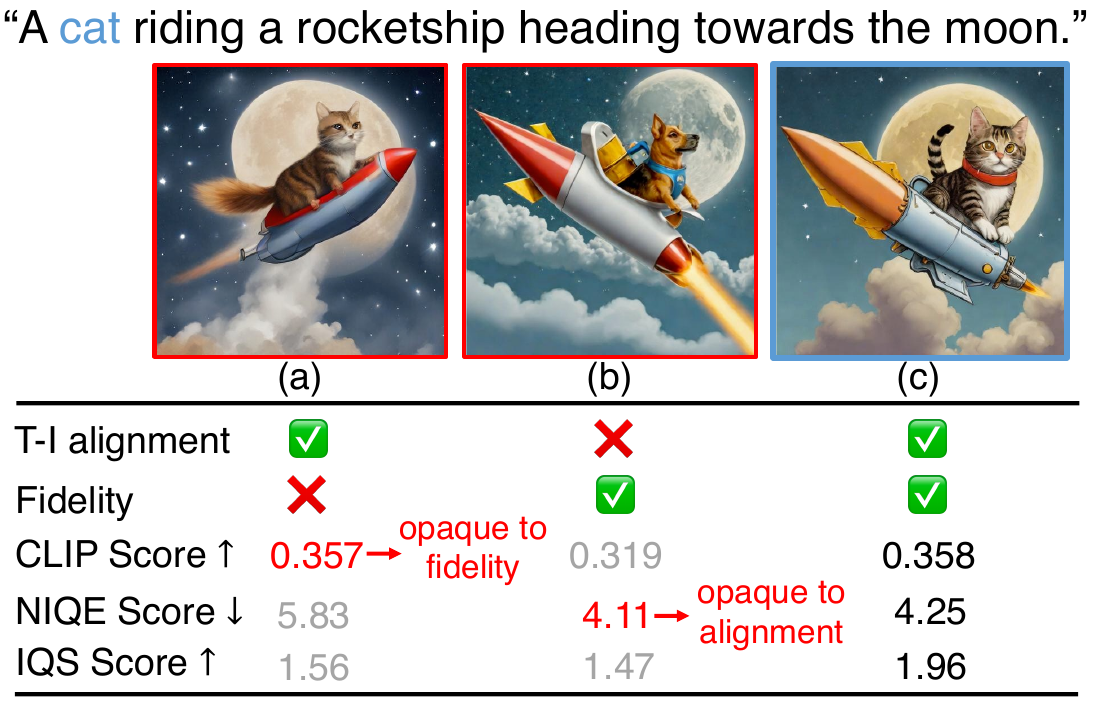}
  \caption{Comparison of image quality evaluation metrics: CLIP Score, NIQE Score, and IQS Score.}
  \label{fig: metric_supp}
\end{figure}

\vspace{0.05in}
\noindent
\textbf{Evaluation Metrics.}
Here, we further discuss the metrics used to evaluate the quality of generated images. Traditional metrics such as CLIP Score and NIQE have certain limitations and lack sufficient objectivity. As shown in Figure~\ref{fig: metric_supp}, a) CLIP Score is opaque to perceptual fidelity and tends to assign high scores to images with high text-image alignment, even if they exhibit noticeable artifacts (such as the moon and paw); b) NIQE is opaque to text-image alignment and tends to assign high scores to images with high perceptual fidelity, even if there is a mismatch between the text and the image (as in the example of the dog). In contrast, the IQS Score comprehensively considers both text-image alignment and high perceptual fidelity, providing a more objective score.

\subsection{More Results}

\begin{table}[h]
\centering
 \ra{1.1}
 \tabcolsep=0.18cm
\scalebox{0.72}{
\begin{tabular}{@{}lccccccc@{}}
\toprule
\multirow{2}{*}{\textbf{Laion-COCO}}      & \multicolumn{2}{c}{\textbf{Speed}} & \multicolumn{5}{c}{\textbf{Image Quality}}    \\ \cmidrule(l){2-3} \cmidrule(l){4-8} 
                 &Step$\downarrow$             &Time$\downarrow$            &IQS$\uparrow$ &CLIP$\uparrow$  &IS$\uparrow$ &FID$\downarrow$   &NIQE$\downarrow$  \\ \midrule
SD-v2.1  & 50                  &    2.27             &0.350	&0.319	&30.71	&22.08	&4.58  \\\midrule
SD-v2.1 & 31                 &    1.41           &0.316	&0.320	&30.01	&22.97	&4.67      \\
Random   & 30.07        &     1.46            &0.291	&0.318	&29.51	&22.45	&4.62      \\
Heuristic                   & 32.75             & 1.58            & 0.315  & 0.320  & 30.21 & 22.25 & 4.60  \\
Perplexity                  & 33.79             & 1.62            & 0.319 & 0.319 & 30.12 & 22.19 & 4.61 \\ \midrule
\rowcolor[HTML]{e9e9e9} 
AdaDiff    & 31.34       & \res{1.50}{$\uparrow$33.9\%}               &\textbf{0.345}	&\textbf{0.321}	&\textbf{30.51}	&\textbf{22.10}	&\textbf{4.58}           \\\bottomrule 
\end{tabular}}
 \caption{Comparison of AdaDiff with other baselines on Laion-COCO text-to-image benchmark.}
\label{tab: results on Laion-COCO}
\end{table}

\vspace{0.05in}
\noindent
\textbf{Performance on Image Generation.}
Table \ref{tab: results on Laion-COCO} offers a comprehensive evaluation of AdaDiff's performance on the Laion-COCO benchmarks. AdaDiff utilizes an average of 31.34 sampling steps. Compared to the fixed 50-step SD, AdaDiff sustains comparable performance across five image quality metrics while achieving a 33.9\% increase in speed. At comparable speeds, AdaDiff significantly surpasses the fixed 31-step SD on all five metrics. Furthermore, the adaptive step usage policy developed by AdaDiff demonstrates clear advantages over random policies and manually crafted strategies such as heuristic and perplexity-based methods. These results confirm that AdaDiff effectively generates instance-specific step usage policies, dynamically adjusting sampling steps per prompt to enhance speed with minimal impact on image quality.

\begin{table}[h]
\centering
 \ra{1.1}
 \tabcolsep=0.18cm
\scalebox{0.72}{
\begin{tabular}{@{}lccccccc@{}}
\toprule
\multirow{2}{*}{\textbf{DiffusionDB}}      & \multicolumn{2}{c}{\textbf{Speed}} & \multicolumn{5}{c}{\textbf{Image Quality}}    \\ \cmidrule(l){2-3} \cmidrule(l){4-8} 
                 &Step$\downarrow$             &Time$\downarrow$            &IQS$\uparrow$ &CLIP$\uparrow$  &IS$\uparrow$ &FID$\downarrow$   &NIQE$\downarrow$  \\ \midrule
 
SD-v2.1                 & 50    & 2.28 & 0.281 & 0.327 & 15.55 & 8.57 & 4.30  \\ \midrule
SD-v2.1               & 32    & 1.43 & 0.237 & 0.327 & 15.48 & 8.67 & 4.46 \\
Random           & 30.01  & 1.42 & 0.221 & 0.326 & 15.34 & 8.74 & 4.48 \\
Heuristic                    & 34.95             & 1.66            & 0.235 & 0.326 & 15.37 & 8.66 & 4.45 \\
Perplexity                   & 34.21             & 1.63            & 0.241 & 0.327 & 15.35 & 8.69 & 4.43 \\ \midrule
\rowcolor[HTML]{e9e9e9} 
AdaDiff          & 32.38 & \res{1.52}{$\uparrow$33.3\%} & \textbf{0.273} & \textbf{0.328} & \textbf{15.52} & \textbf{8.56} & \textbf{4.33} \\ \bottomrule 
\end{tabular}}
 \caption{Comparison of AdaDiff with other baselines on DiffusionDB text-to-image benchmark.}
\label{tab: results on DiffusionDB}
\end{table} 

Table \ref{tab: results on DiffusionDB} showcases AdaDiff's performance on the DiffusionDB benchmarks, where AdaDiff utilizes an average of 32.38 sampling steps. Compared to the fixed 50-step SD, AdaDiff achieves similar performance across five image quality metrics while offering a 33.3\% improvement in speed. At comparable speeds, AdaDiff significantly outperforms the fixed 32-step SD, as well as random, heuristic, and perplexity policies in terms of image quality. These findings also validate the effectiveness of AdaDiff's instance-specific step usage policies.

\begin{figure*}[h]
  \centering
  \includegraphics[width=1.0\linewidth]{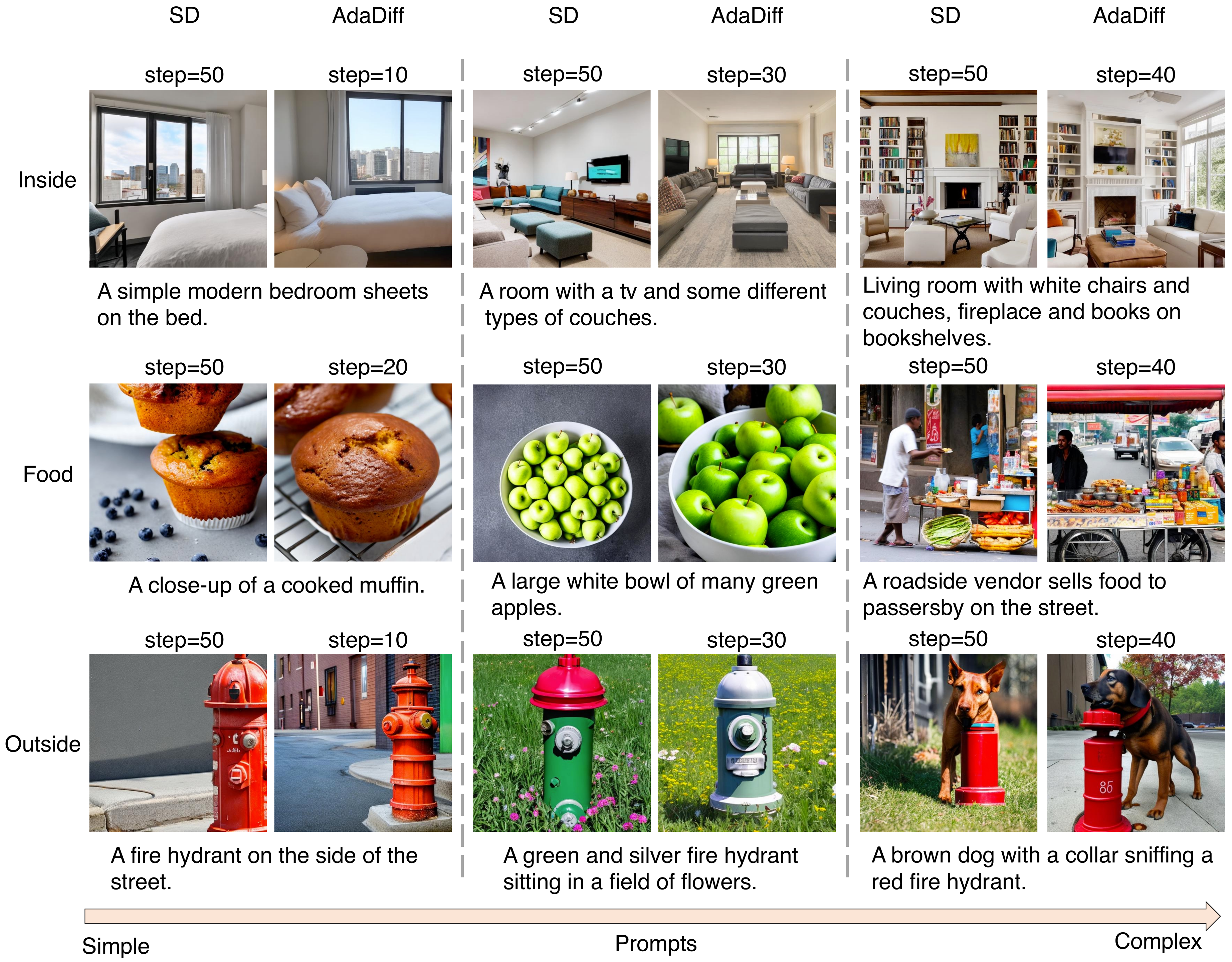}
  \caption{\textbf{Additional Qualitative Results.} We validate the effectiveness of AdaDiff in three scenarios (indoor, food, and outdoor). As the richness of information in the prompts gradually increases from left to right, AdaDiff allocates more reasonable denoising steps through instance-specific step usage policies, achieving accelerated speed while maintaining comparable image quality.}
  \label{fig: visual_supp}
\end{figure*}

\vspace{0.05in}
\noindent
\textbf{More Qualitative Results.}
In Figure~\ref{fig: visual_supp}, we provide additional visualization examples of AdaDiff in text-to-image generation. We validate several common scenarios, such as indoor, food, and outdoor, and choose prompts with varying degrees of information richness from these scenarios (reflected in the number of objects, relationships between objects, diversity of attributes, \etc). The findings in ~\ref{fig: visual_supp} demonstrate that compared to the Stable Diffusion with fixed 50-step generations, AdaDiff allocates sample-specific denoising steps based on the richness of input text, achieving faster generation speed while maintaining high-quality images.

\bibliography{main}

\end{document}